\documentclass[letterpaper, 10 pt, conference]{ieeeconf}  

\IEEEoverridecommandlockouts                              

\usepackage{graphics} 
\usepackage{epsfig} 
\usepackage{mathptmx} 
\usepackage{times} 
\usepackage{amsmath} 
\usepackage{amssymb,bm}  
\usepackage{graphicx}
\usepackage{color}
\usepackage{lipsum}

\usepackage{xcolor} 
\usepackage[backend=biber,style=ieee]{biblatex}

\DeclareSourcemap{
  \maps{
    \map{
      \pertype{article}
      \step[fieldset=language, null]
      \step[fieldset=url, null]
      \step[fieldset=doi, null]
      \step[fieldset=issn, null]
      \step[fieldset=isbn, null]
      \step[fieldset=note, null]
      \step[fieldset=editor, null]
      \step[fieldset=urldate, null]
      \step[fieldset=file, null]
    }
  }
}
\DeclareSourcemap{
  \maps{
    \map{
      \pertype{inproceedings}
      \step[fieldset=language, null]
      \step[fieldset=url, null]
      \step[fieldset=doi, null]
      \step[fieldset=issn, null]
      \step[fieldset=isbn, null]
      \step[fieldset=note, null]
      \step[fieldset=editor, null]
      \step[fieldset=urldate, null]
      \step[fieldset=file, null]
    }
  }
}
\DeclareSourcemap{
  \maps{
    \map{
      \pertype{incollection}
      \step[fieldset=language, null]
      \step[fieldset=url, null]
      \step[fieldset=doi, null]
      \step[fieldset=issn, null]
      \step[fieldset=isbn, null]
      \step[fieldset=note, null]
      \step[fieldset=editor, null]
      \step[fieldset=urldate, null]
      \step[fieldset=file, null]
    }
  }
}

\title{\LARGE \bf
Conjugate Momentum-Based Estimation of External Forces for Bio-Inspired Morphing Wing Flight
}

\author{Bibek Gupta$^{1}$, Eric Sihite$^{2}$, and Alireza Ramezani$^{1*}$
\thanks{$^{1}$Authors are with the Silicon Synapse Labs, Department of
Electrical and Computer Engineering, Northeastern University, Boston,
USA. Emails: gupta.bi, a.ramezani@northeastern.edu}%
\thanks{$^{2}$Author is with the Department of Aerospace Engineering,
California Institute of Technology, Pasadena, USA. Email:
esihite@caltech.edu}%
\thanks{*Corresponding author.}
}

\begin{document}

\maketitle
\thispagestyle{empty}
\pagestyle{empty}

\begin{abstract}

Dynamic morphing wing flights present significant challenges in accurately estimating external forces due to complex interactions between aerodynamics, rapid wing movements, and external disturbances.  Traditional force estimation methods often struggle with unpredictable disturbances like wind gusts or unmodeled impacts that can destabilize flight in real-world scenarios. This paper addresses these challenges by implementing a Conjugate Momentum-based Observer, which effectively estimates and manages unknown external forces acting on the Aerobat, a bio-inspired robotic platform with dynamically morphing wings. Through simulations, the observer demonstrates its capability to accurately detect and quantify external forces, even in the presence of Gaussian noise and abrupt impulse inputs. The results validate the robustness of the method, showing improved stability and control of the Aerobat in dynamic environments. This research contributes to advancements in bio-inspired robotics by enhancing force estimation for flapping-wing systems, with potential applications in autonomous aerial navigation and robust flight control.

\end{abstract}

\section{Introduction}

The precision with which biological entities navigate and manipulate their environments is often highlighted by their ability to dynamically respond to external stimuli. This capability is crucial, as demonstrated by numerous studies on various species. For instance, locusts in flight utilize abdominal movements to adjust their body orientation in response to external disturbances \cite{bomphrey_tomographic_2012}. Similarly, vertebrates like lizards manipulate their tails to maintain or change orientation \cite{libby_tail-assisted_2012}, while bats employ complex wing movements to perform intricate aerial maneuvers \cite{riskin_bats_2009,ramezani_aerobat_2022,ramezani_biomimetic_2017,hoff_optimizing_2018,sihite_wake-based_2022,sihite_morphology-centered_2024,gupta_banking_2024}.

In contrast to these biological systems, our research focuses on the robotic platform Aerobat, which is designed to replicate bat-like flight dynamics. Aerobat features large, dynamically morphing wings that pose significant challenges for modeling and control due to their rapid flexion and expansion \cite{sihite_wake-based_2022,sihite_computational_2020,ramezani_bat_2016}. While we have already developed a robust aerodynamic model \cite{sihite_unsteady_2022} that accurately estimates aerodynamic forces, our work goes beyond traditional force estimation approaches that primarily address aerodynamic and inertial forces \cite{karasek_free_2016}. Instead, we focus on estimating unknown external forces—such as wind gusts or mechanical impacts—which are often unaccounted for in standard models.

External disturbances, such as wind gusts or sudden impacts, pose significant challenges to the stability and control of flapping-wing robots. The unsteady and highly nonlinear aerodynamics inherent in flapping flight make these systems particularly sensitive to external perturbations \cite{ho_unsteady_2003}. Wind gusts can induce significant changes in aerodynamic forces and moments, leading to the vehicle's loss of control or destabilization \cite{bhatia_stabilization_2014}. Additionally, sudden impacts or collisions can disrupt the delicate balance required for stable flapping flight, potentially causing mechanical damage or loss of flight capability \cite{keennon_development_2012}. Detecting these external forces is crucial for understanding their effects and developing strategies to mitigate their impact on flight stability.

To address this challenge, we employ a Conjugate Momentum-based Observer, a sophisticated tool traditionally used in legged robotics to detect external forces \cite{vorndamme_collision_2017}. This observer's application to flapping-wing robotics represents a novel approach, especially given the intricate dynamics of systems like the Aerobat. Theoretical frameworks for such observers are well-established \cite{albuschaffer_dlr_2007}, yet their practical application to flapping-wing flight remains relatively unexplored.

Our study aims to demonstrate the effectiveness of using a Conjugate Momentum-based Observer to estimate unknown external forces on the Aerobat. This methodology enhances the understanding of how external forces impact bio-inspired robotic flyers while improving the robustness and reliability of their control systems in unpredictable environments. This research contributes significantly to the field of robotic flight, particularly in scenarios where external disturbances play a critical role.

\begin{figure}
    \centering
    \includegraphics[width=0.9\linewidth]{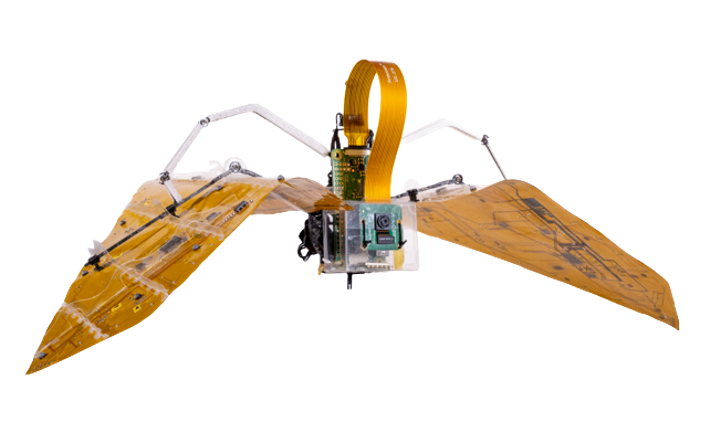}
    \caption{Shows Aerobat platform \cite{sihite_actuation_2023}. The platform is designed to study inertial and aerodynamic dynamics' contribution roles in dynamic morphing wing flight.}
    \label{fig:cover-image}
\end{figure}

This work is organized as follows. Section \ref{sec:model} details the kinematics, dynamics, and aerodynamics modeling of the Aerobat platform. In Section \ref{sec:estim}, we elaborate on the theoretical framework and the implementation of the Conjugate Momentum-based Observer used for estimating external forces. Section \ref{sec:res} presents a comprehensive analysis of the simulation setup, methodology, and results demonstrating the observer's effectiveness in real-world scenarios. Finally, Section \ref{sec:concl} discusses the implications of these findings for the field of bio-inspired robotics and outlines potential avenues for future research.

\section{Modeling}
\label{sec:model}

\begin{figure}
    \centering
    \includegraphics[width=0.9\linewidth]{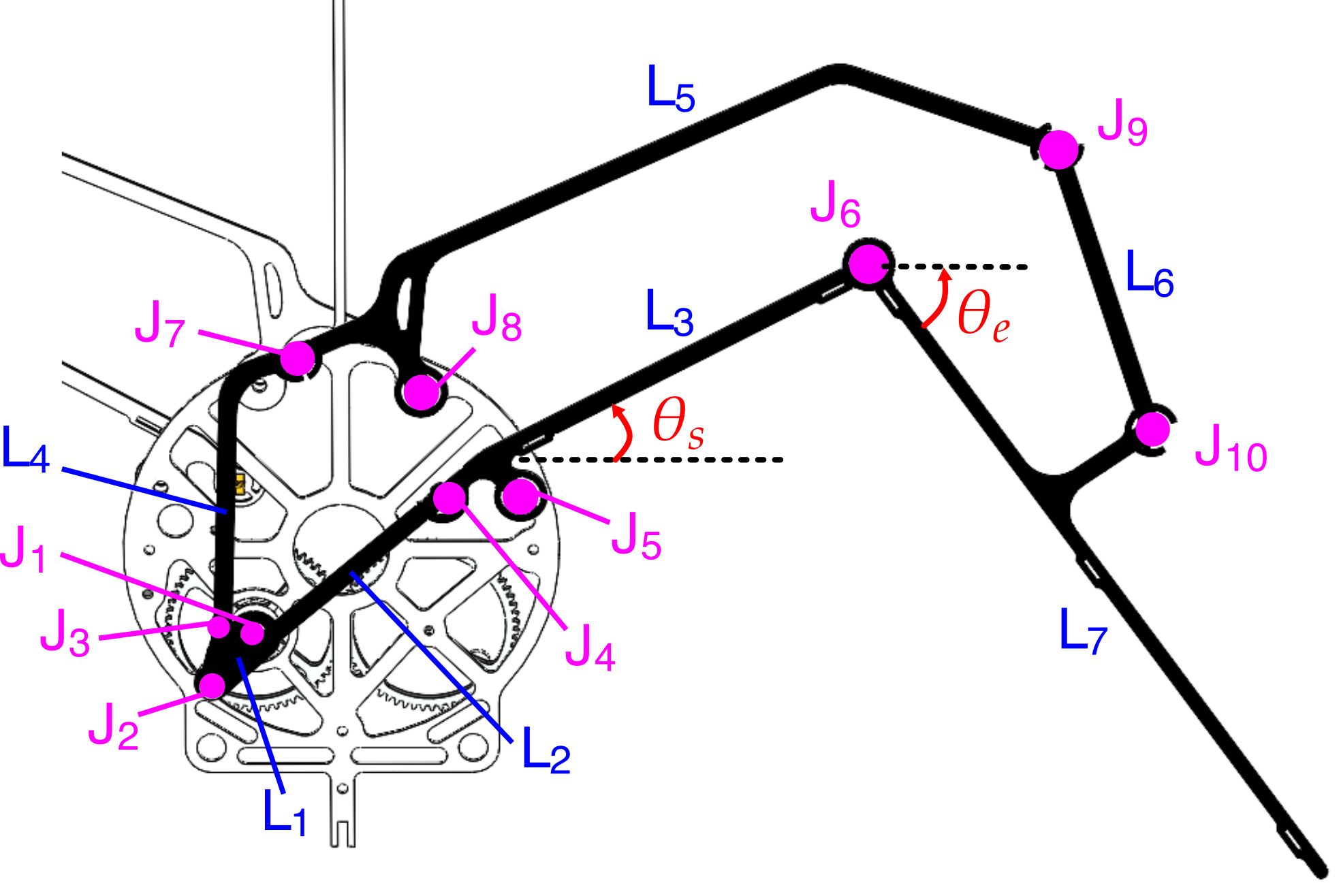}
    \caption{Schematic of the Kinetic Sculpture (KS) mechanism featuring 7 linkages (L1 to L7) and 10 joints (J1 to J10). The KS is uniquely engineered to be driven by a single actuator. This design includes two pivotal angles: $\theta_s$ and $\theta_e$ representing the shoulder and elbow joint angles respectively \cite{sihite_actuation_2023}.}
    \label{fig:ks}
\end{figure}

The Aerobat system, as shown in Fig.~\ref{fig:cover-image}, 
 can be described using five distinct rotating bodies connected by joints or hinges, illustrated in Figs.~\ref{fig:ks} and \ref{fig:aerobat_model}. These include the main body, along with the proximal and distal segments of both left and right wings. This arm-wing mechanism is designed to imitate the natural degrees of freedom (DoF) of a bat's wing, as depicted in Fig.~\ref{fig:ks}. The degrees of freedom in Aerobat’s flapping wing closely resemble the natural movement of a bat's wings. The shoulder joint plunge angle ($\theta_s$) plays a key role in controlling the upstroke and downstroke of the wings, forming the primary flapping motion essential for flight.
The elbow flexion and extension angle ($\theta_e$) allows the wing to expand during the downstroke and retract during the upstroke. This movement enhances the wing's efficiency by reducing negative lift during the upstroke, thereby optimizing the overall flight performance. 

To simplify the dynamic modeling of the complex Aerobat system, several assumptions are made: the left and right-wing linkages are treated as coupled, allowing their movements to be coordinated; the linkages of the kinetic sculpture (KS) are assumed to be massless; and KS kinematic constraints are enforced.

\subsection{KS Kinematic Constraints Modeling}

The Kinetic Sculpture (KS) system, illustrated in Fig.~\ref{fig:ks}, is modeled as a constrained planar linkage system with 10 joints and 7 linkages. Each joint, denoted $i$, is defined by an angle $ {q}_i$ and a position $ {p}_i \in \mathbb{R}^2$. The motion dynamics are captured by kinematic equations listed below, which illustrate the interdependencies within the linkages.
\begin{equation}
  \begin{aligned}  
    \ddot{p}^1_{4} = \ddot{p}^5_{4}, \quad \ddot{p}^1_{7} = \ddot{p}^8_{7}, \quad \ddot{p}^5_{10} = \ddot{p}^8_{10}, \quad \ddot{q}_1 = u_k,
  \end{aligned}
\end{equation}

The system's primary degree of freedom (DOF), $\ddot{ {q}}_1$, is governed by the motor's angular acceleration $u_k$. This simplifies to a set of equations that model the joint accelerations essential for the dynamic behavior of the structure: $\ddot{ {q}}_k = [\ddot{ {q}}_1, \ddot{ {q}}_2, \ldots, \ddot{ {q}}_9]$.
Let state vectors $ {x}_k = [ {q}_k, \dot{ {q}}_k]$ define the kinematic subsystem:
\begin{equation}
  \begin{aligned}
    \Sigma_{K}&:\left\{
  \begin{aligned}
      \dot{x}_k &= f_k(x_k) + g_k(x_k) u_k, \\
       y_k &= [\ddot{q}_6, \ddot{q}_7] = C_k (f_k(x_k) + g_k(x_k) u_k),
   \end{aligned}
    \right.
   \end{aligned}
\label{eq:kinematic}
\end{equation}

Here, $ {y}_k$ denotes the accelerations at the shoulder and elbow joints, functioning as dynamic constraints that are effectively mapped to $u_m$, a generalized motor torque. This torque acts as an input to the Aerobat's dynamic model, facilitating precise control over the joints' movements.

\subsection{Flight Dynamics Modeling}
\begin{figure}
    \centering
    \includegraphics[width=0.9\linewidth]{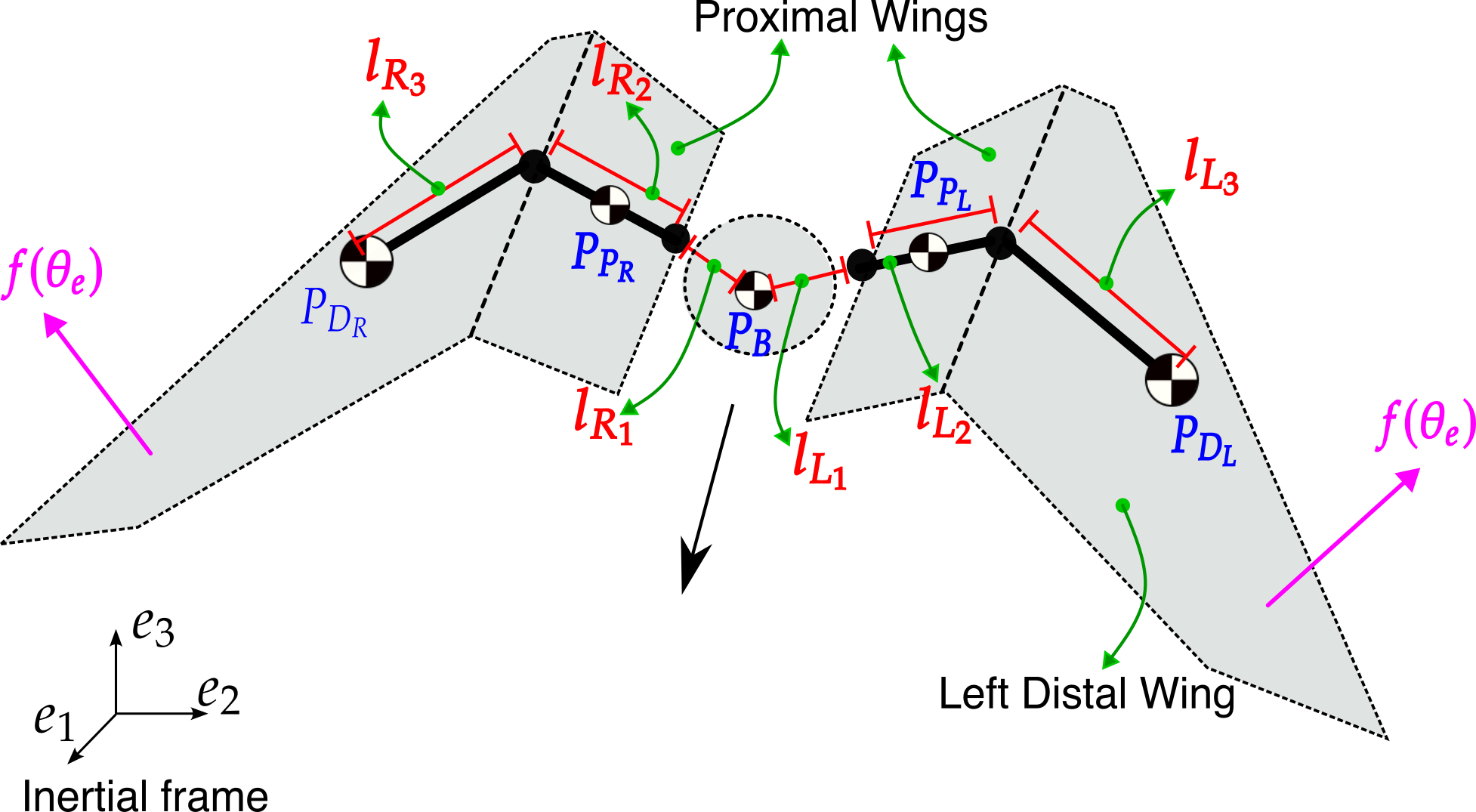}
    \caption{Aerobat with simplified linkages model using five rotating bodies, the linear position $ {p}$ represents the center of mass, and $ {l}$ represents the length vectors relevant to the Aerobat model conformation.}
    \label{fig:aerobat_model}
    \vspace{-0.05in}
\end{figure}

In this context, vectors with superscripts are defined in non-inertial coordinate frames, while those without superscripts are defined in the inertial frame. For example, $ {x}^B$ represents the vector $ {x}$ relative to frame B. The rotations between the reference frames for the five bodies can be described as follows:
\begin{equation}
    \begin{aligned}
         {x} &= R_B\, {x}^{B}, \quad &
         {x}^B &= R_{P_L}\, {x}^{P_L} = R_{P_R}\, {x}^{P_R} \\
         {x}^{P_L} &= R_{D_L}\, {x}^{D_L}, \quad &
         {x}^{P_R} &= R_{D_R}\, {x}^{D_R},
    \end{aligned}
\label{eq:rotation_matrices}
\end{equation}

In these equations, ${R}_B$ represents the rotation matrix of the body relative to the inertial frame. ${R}_{P_L}$ and ${R}_{P_R}$ are the rotation matrices for the left and right proximal wings with respect to the body frame. Additionally, ${R}_{D_L}$ and ${R}_{D_R}$ correspond to the rotations of the left and right distal wings relative to their respective arms.

The angular velocities associated with these rotation matrices can also be expressed within their specific coordinate frames. This leads to the following pairs of rotation matrices and angular velocities:
$( {R}_B, {\omega}_B)$, $( {R}_{P_L},  {\omega}^B_{P_L})$, $( {R}_{P_R},  {\omega}^B_{P_R})$, $( {R}_{D_L},  {\omega}^B_{D_L})$, and $( {R}_{D_R},  {\omega}^B_{D_R})$.

Let the angles $\theta_s$ and $\theta_e$ denote the shoulder and elbow angles, respectively, which are biologically relevant flapping angles, as illustrated in Fig.~\ref{fig:ks}. Here the superscripts $L$ and $R$ signify the left and right wings. The rotation matrices for the left proximal and distal wings can then be expressed as follows:

\begin{equation}
    \begin{aligned}
         {R}_{P_L} =   {R}_x(\theta^L_s), \quad  {R}_{D_L} =  {R}_x(\theta^L_e)
    \end{aligned}
\end{equation}
Here, $ {R}_x(\theta)$ represents the rotation matrices about the x-axis. The angular velocities for the left proximal-distal wings are defined as follows:
\begin{equation}
    \begin{aligned}
         {\omega}_{P_L}^{B} &= \begin{bmatrix} \dot{\theta}_p^L, 0 , 0 \end{bmatrix}^\top +  {\omega}_{B}^{B}\\
         {\omega}_{D_L}^{P_L} &= \begin{bmatrix} 
        \dot{\theta}_e^L, 0 , 0 \end{bmatrix}^\top +  {\omega}_{P_L}^{P_L}.
    \end{aligned}
\label{eq:angular_vel_left}
\end{equation}

Based on our assumption regarding the coupling of the right and left wings, the subsequent sections of this paper will concentrate solely on the left wings. The derivations for the right wings can be similarly formulated.

As illustrated in Fig.~\ref{fig:aerobat_model}, let $ {p}^B$ represent the linear position of the center of mass of a body, while $ {l}^L_j$ and $ {l}^R_j$ (where $j = \{1, 2, 3\}$) denote the length vectors that describe the morphology of the Aerobat mechanism. These length vectors are constant relative to their local frame of reference. The linear position of the center of mass for the left proximal and distal wings can be expressed as follows:

\begin{equation}
    \begin{aligned}
         {p}_{P_L} &=  {p}_B + R_{B}\, {l}_{L1}^{B} + \tfrac{1}{2}\,R_B\,R_{P_L}\, {l}_{L2}^{P_L}
         \\
         {p}_{D_L} &=  {p}_{P_L} + \tfrac{1}{2}\,R_B\,R_{P_L}\, {l}_{L2}^{P_L} + R_B\,R_{P_L}\,R_{D_L}\, {l}_{L3}^{D_L},
    \end{aligned}
\label{eq:linear_position}
\end{equation}

The linear velocity of the center of mass can be obtained by differentiating the linear positions from \eqref{eq:linear_position} with respect to time.

The kinetic and potential energy of the system can be derived as follows:
\begin{equation}
    \begin{aligned}
    T &= \sum_{F \in \mathcal{F}} \left( m_F\,\dot{ {p}}_F^\top\,\dot{ {p}}_F + ( {\omega}_F^{F})^\top\,\hat{I}_{F}\, {\omega}_F^{F} \right)\frac{1}{2} \\
    U &= \sum_{F \in \mathcal{F}} m_F \, [ 0, 0, g ] \,  {p}_F,
    \end{aligned}
\label{eq:energy}
\end{equation}

where $\mathcal{F} = \{B, P_L, P_R, D_L, D_R\}$ represents the set of reference frames. In this context, $m_F$ denotes the mass and $\hat{ {I}}_F$ signifies the inertia matrix of the corresponding body. The inertia matrix $\hat{ {I}}_F$ is defined with respect to the local frame of reference and is both diagonal and constant. 

Let $ {q}_d = [ {p}^\top_B, q_s, q_e]$ represent the states of the system, where $ {p}^B \in \mathbb{R}^3$ denotes the position of the center of mass in the inertial frame, and $q_s$ and $q_e$ are the shoulder and elbow joint angles, respectively. Additionally, let $ {\omega}^B \in \mathbb{R}^3$ represent the angular velocity of the body, which can be expressed as $\dot{ {R}}_B =  {R}_B[ {\omega}^B]_\times$, where $[\cdot]_\times$ denotes the skew-symmetric operator.

The equations of motion for the remaining states can be derived by applying the Euler-Lagrange equation. This equation provides a systematic way to obtain the dynamics of a system, where the Lagrangian $L$ is given by $L = T - U$ from \eqref{eq:energy}. The Euler-Lagrange equation for each generalized coordinate $q_i$ is expressed as:

\[
\frac{d}{dt} \left( \frac{\partial L}{\partial \dot{q}_d} \right) - \frac{\partial L}{\partial q_d} = u_d.
\]
where $ {u}_{d}$ is the non-conservative force about the generalized coordinate $ {q}_{d}$.

Finally, let $ {a}_d = [\ddot{ {q}}_d, \dot{ {\omega}}^B]$ represent the acceleration vector of the dynamic states. 
By solving the Euler-Lagrange equations of motion, we obtain the following constrained equation of motion:
\begin{equation}
    \begin{aligned}
     {M}_d( {q}_d,  {R}_B)  {a}_d =  {h}_d( {q}_d, \dot{ {q}}_d,  {R}_B,  {\omega}^B) +  {u}_a +  {u}_m +  {u}_f,
    \end{aligned}
\label{eq:eom}
\end{equation}

Here, $ {M}_d$ represents the mass and inertia matrix, while $ {h}_d$ is a vector that includes the Coriolis and gravitational terms. The term $ {u}_a$ refers to the generalized aerodynamic forces acting on the system, and $ {u}_m$ is the generalized motor torque acting on the wing joints which is selected to directly actuate the joints angles $ {q}_s$ and $ {q}_e$. $ {u}_f$ is some unknown generalized force form of $f(\theta_e)$ acting on a point location on the wing as shown in Fig.~\ref{fig:aerobat_model}.

\subsection{Aerodynamic Modeling}
The aerodynamic behavior of the Aerobat in flight can be captured through the use of an indicial model \cite{sihite_unsteady_2022}. The system is described as:

\begin{equation}
  \begin{aligned}
    \Sigma_{Aero}&:\left\{
  \begin{aligned}
     \dot{\xi} &= A_\xi(t) \xi + B_\xi(t) y_1 \\
      u_a &= C_\xi(t) \xi + D_\xi(t) y_1
   \end{aligned}
    \right.
   \end{aligned}
\label{eq:aero-fulldyn}
\end{equation}

In this formulation, $\xi$ refers to the aerodynamic state variables, while $u$ represents the aerodynamic force, providing insight into the external forces acting on the wings during flapping flight. As shown in Eq.~\ref{eq:aero-fulldyn}, the aerodynamic system is modeled in state-space form, where the matrices $A_\xi$, $B_\xi$, $C_\xi$, and $D_\xi$ govern the aerodynamic response over time \cite{sihite_unsteady_2022}.
This state-space formulation is based on Wagner's indicial model and Prandtl’s lifting line theory, both of which are fundamental to fluid dynamics \cite{sihite_unsteady_2022}.

The indicial model is advantageous because it allows for the efficient calculation of unsteady aerodynamic forces and moments, which is crucial for optimization-based control strategies in flying systems. Furthermore, it can predict wake structures that result from vortex shedding, which is essential for describing flight gaits, similar to those observed in biological systems such as bat flight \cite{hubel_wake_2010,parslew_theoretical_2013}.

Knowing $ {u}_m$ from the kinematic constraint model and $ {u}_a$ from the aerodynamic model, we are left with the unknown term $ {u}_f$. The next step is to estimate this unknown force using a conjugate momentum-based observer to complete our Aerobat model. For simplicity, we assume that the unknown external force acts as a point force on each wing as shown in Fig.~\ref{fig:aerobat_model}.

\section{Conjugate Momentum Based Force Estimation}
\label{sec:estim}

\begin{figure}
    \centering
    \includegraphics[width=0.95\linewidth]{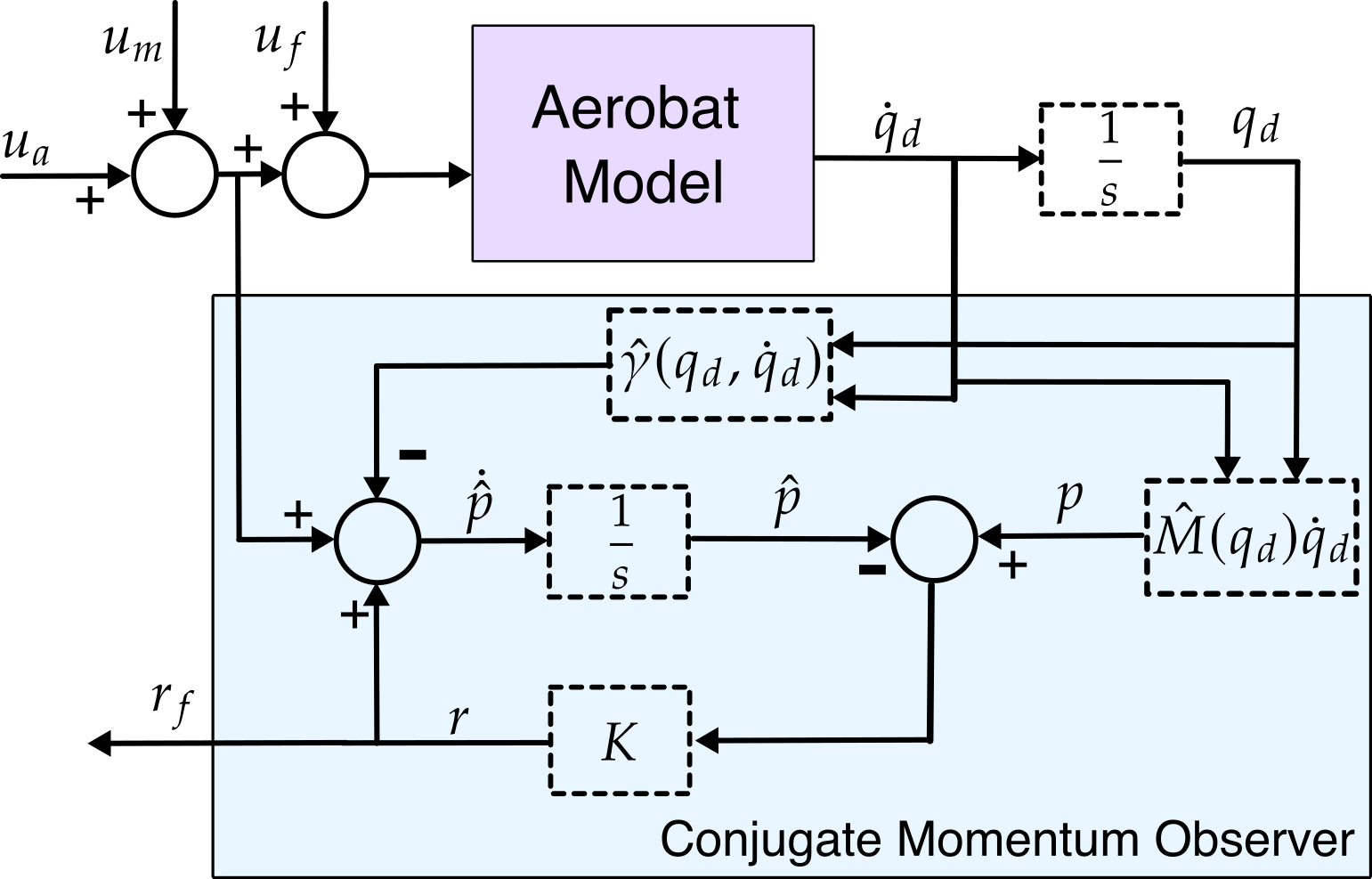}
    \caption{Block diagram of the conjugate momentum observer}
    \label{fig:observer_diag}
\end{figure}

In this section, we focus on estimating external forces acting on specific points of the Aerobat's wings, as shown in Fig~\ref{fig:aerobat_model}. The estimation approach illustrated in Fig.~\ref{fig:observer_diag}, is based on the conjugate momentum observer introduced in \cite{de_luca_sensorless_2005,de_luca_collision_2006,haddadin_robot_2017}, a widely recognized technique in dynamic systems control. This observer enables the estimation of unmeasured external forces, such as disturbances or unknown inputs, using the system's state information and known inputs.

The conjugate momentum of the system, denoted as $ {p}(t)$, is defined as the product of the mass-inertia matrix and the generalized velocity:
\begin{equation}
\begin{aligned}
 {p}(t) =  {M}_d { \dot{q}}_d
\end{aligned}
\end{equation}

This variable encapsulates critical information about the system's dynamics and is influenced by both internal and external forces.

To estimate the external forces, we construct a signal $ {r}(t)$, that evolves according to the equation:

\begin{equation}
\begin{aligned}
\dot{r} = Kr + K {u}_f 
\end{aligned}
\label{eq:signal}
\end{equation}

This equation represents $ {r}(t)$ as a low-pass filtered version of the external forces acting on the Aerobat \cite{de_luca_actuator_2003}. To demonstrate this relationship, we utilize the following established equation:
\begin{equation}
  \begin{aligned}
    \dot{M} = C( {q_d}, \dot{q}_d) + C^T( {q_d}, \dot{q}_d)
    \\
    \gamma({q_d}, \dot{q}_d) = G( {q_d}) - C^T( {q_d}, \dot{q}_d){ \dot{q}_d}
  \end{aligned}
\label{eq:property}
\end{equation}

By integrating this signal Eq.~\eqref{eq:signal} over time, we derive an estimate of the unknown external forces:
\begin{equation}
r_f(t) = K \left( p(t) - \int_0^t \left( {u}_a +  {u}_m - \hat{\gamma}( {q_d}, \dot{q}_d) + r(s)\right) ds - p(0) \right)
\end{equation}
where $K$ is a fixed diagonal gain matrix, and $p(t)$ is the conjugate momentum.This method provides a robust framework for estimating external forces, such as unmodeled disturbances, during the Aerobat's flight.

The conjugate momentum observer thus plays a crucial role in real-time force estimation, helping to understand the external dynamics that impact the Aerobat during its flight.

\subsection{Modeling of Temporal Noise and Step Input in Force Estimation}

In our simulation, accurately modeling external disturbances, such as noise and transient inputs, is critical for assessing the Aerobat’s dynamics. We introduce a Gaussian noise component to simulate environmental variability and sensor noise. This noise is characterized by a mean $\mu = 0$ and a standard deviation $\sigma = 0.01$, and is mathematically represented in the force model as follows:

\begin{equation}
\begin{aligned}
\text{noise} \sim \mathcal{N}(0, 0.01^2)
\end{aligned}
\label{eq:noise}
\end{equation}

This noise simulates realistic operational conditions and challenges the system's ability to estimate forces accurately.

Additionally, a step input is integrated into the force calculations to simulate sudden changes in external forces, which can occur due to abrupt maneuvers or environmental disturbances. The force $f_{t}$ at each timestep is calculated using the following conditional equation:

\begin{equation}
    \begin{aligned}
        f_{t} = 
        \begin{cases} 
            0.2 \sin(q_e) + \text{noise} + 0.15, & \text{if } 1 \text{ sec} < t \leq 1.6 \text{ sec} \\
            0.2 \sin(q_e) + \text{noise}, & \text{otherwise}
\end{cases}
\end{aligned}
\end{equation}

Here, $q_e$ represents the elbow joint angle, and the step input of 0.15 is applied within a specified time window during the simulation. This implementation allows for the study of the system’s response to abrupt perturbations, which is crucial for evaluating the robustness of control and estimation algorithms.

By incorporating both Gaussian noise and step inputs, we enhance the realism of the simulation, making the model more representative of actual flight conditions. This approach enables a comprehensive evaluation of the Aerobat’s dynamic performance and resilience to external disturbances.

\section{Results}
\label{sec:res}
\begin{figure}[t]
    \centering
    \includegraphics[width=0.95\linewidth]{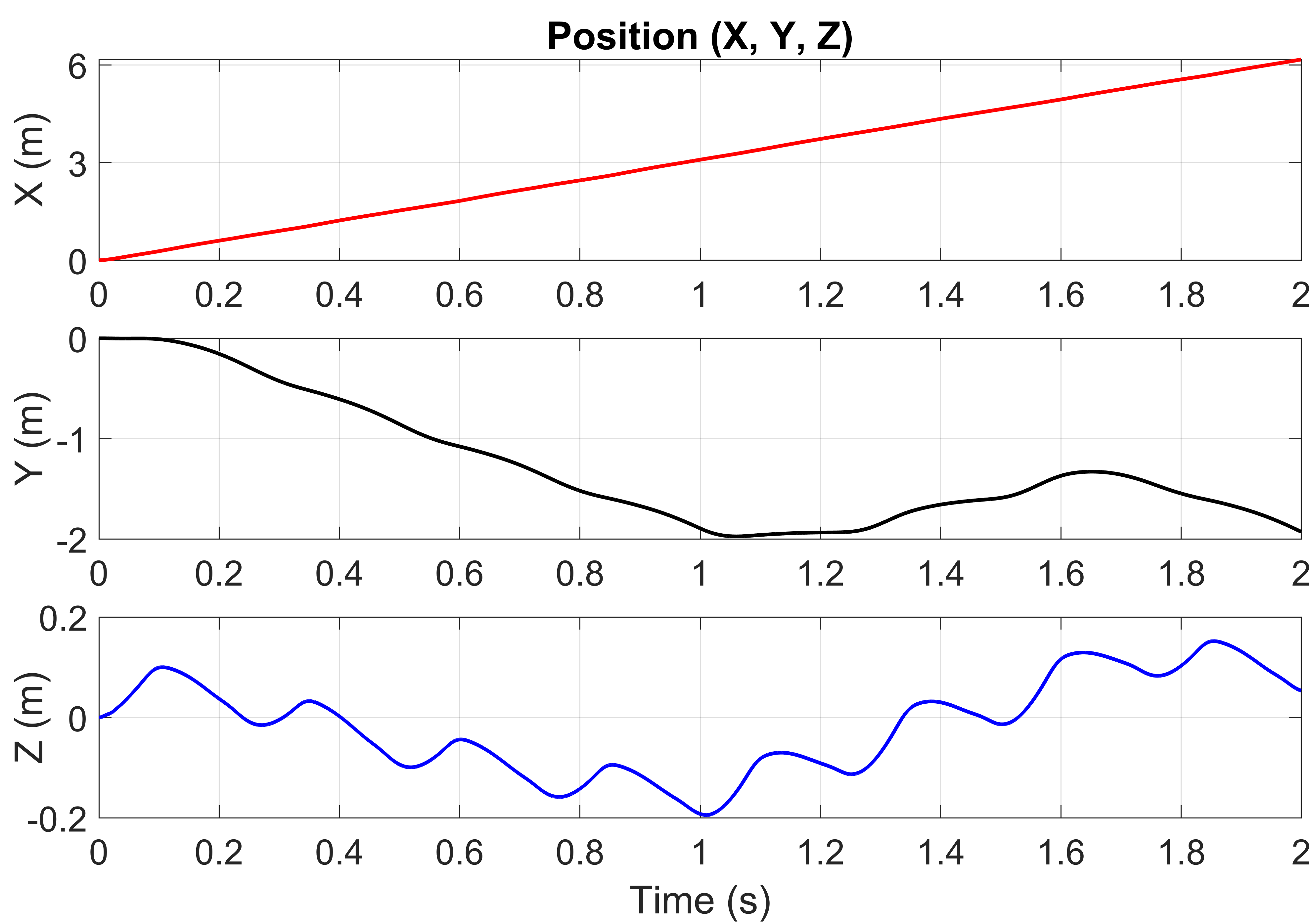}
    \caption{Shows position of Aerobat in a 3D space over a 2-second interval.}
    \label{fig:position}
\end{figure}
\begin{figure}[t]
    \centering
    \includegraphics[width=0.95\linewidth]{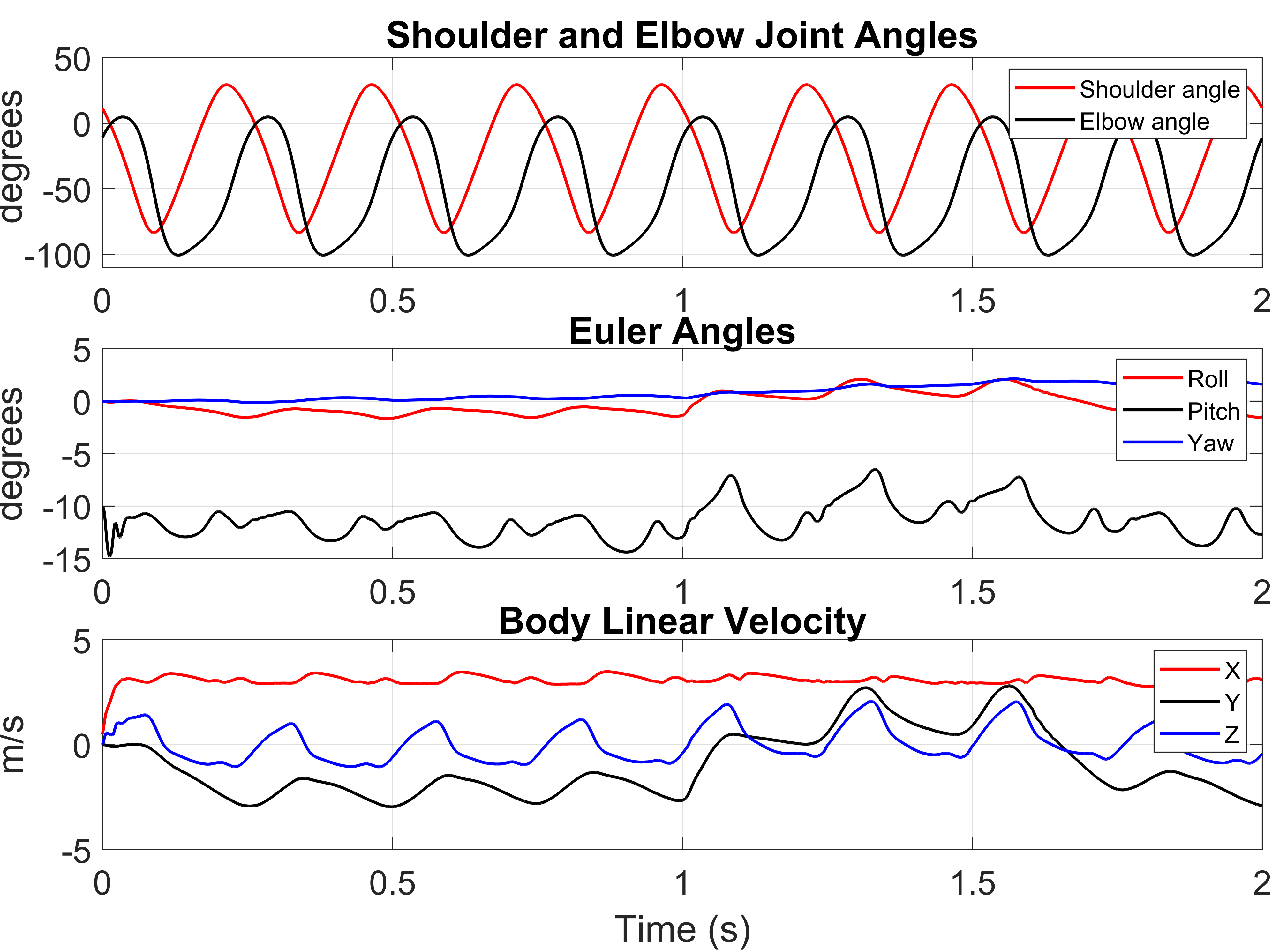}
    \caption{Dynamics of Aerobat depicting shoulder and elbow joint angles (top), Body Euler angles (roll, pitch, yaw) over time (middle), and the three-dimensional linear velocities (bottom).}
    \label{fig:states}
\end{figure}
\begin{figure}[t]
    \centering
    \includegraphics[width=0.95\linewidth]{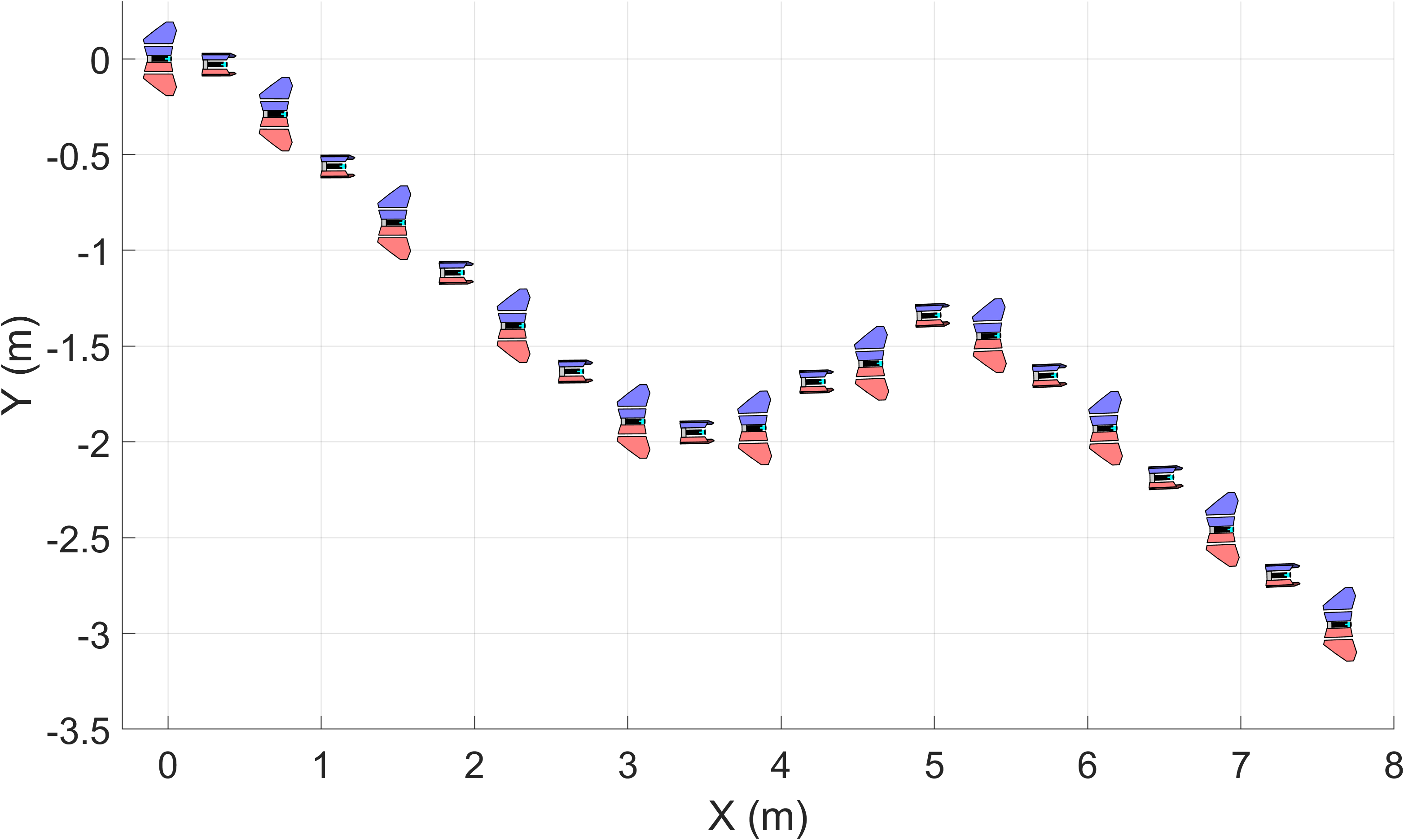}
    \caption{Top view snapshots of the simulation showing the trajectory of the Aerobat.}
    \label{fig:stick-diagram}
\end{figure}
\begin{figure}[t]
    \centering
    \includegraphics[width=0.95\linewidth]{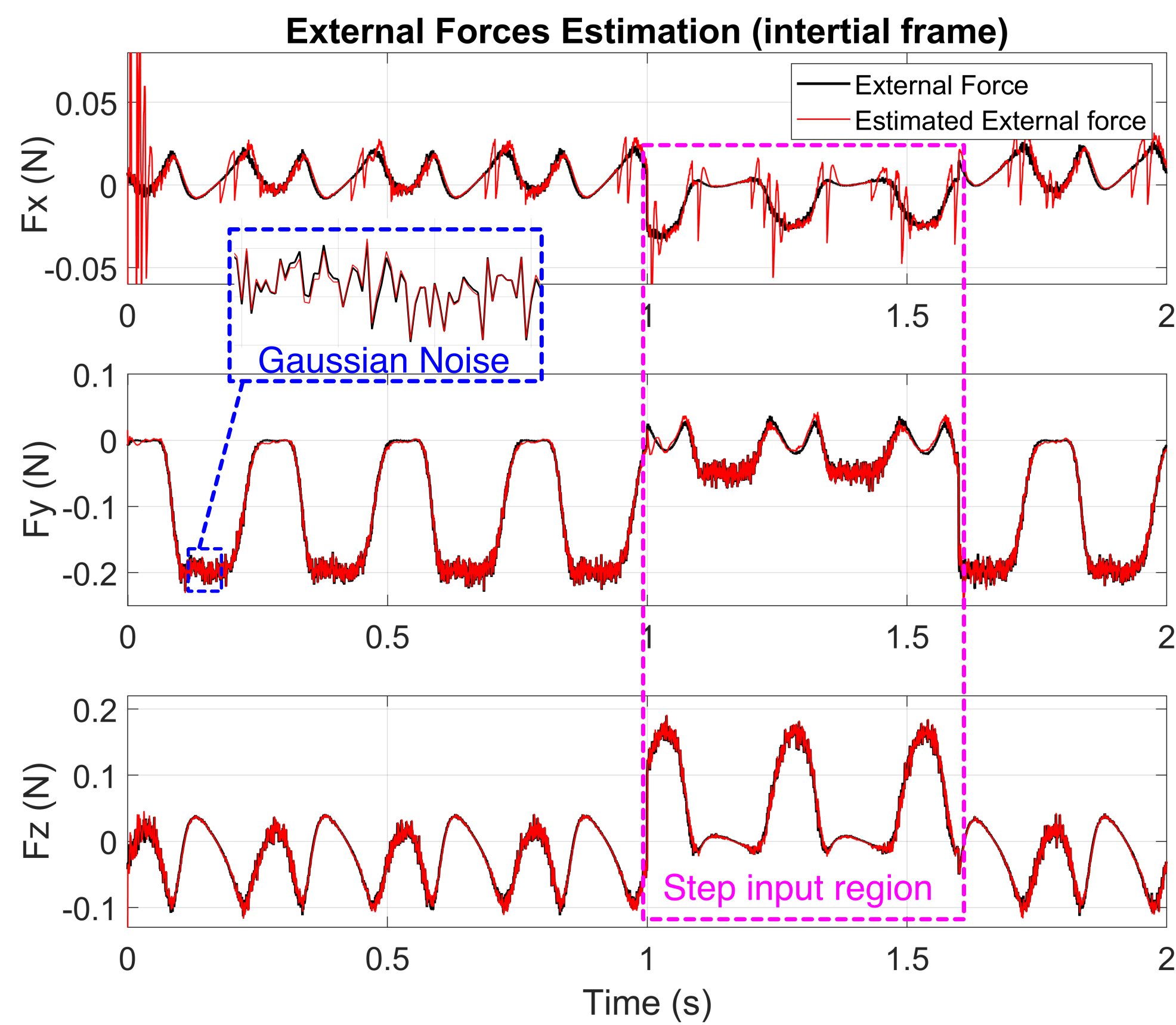}
    \caption{Estimation of unknown external forces acting on Aerobat's wing, compared with the actual input forces.}
    \label{fig:estim-diagram}
\end{figure}

The numerical simulation for estimating unknown external forces acting on the Aerobat was conducted in MATLAB, utilizing the dynamic equations, kinematic constraints, and aerodynamic model detailed in section \ref{sec:model} and \ref{sec:estim}. The fourth-order Runge-Kutta method was employed to ensure high accuracy in integrating the system’s differential equations. The model's effectiveness was evaluated by simulating various wing morphologies under different flight conditions, focusing on validating the conjugate momentum-based observer's ability to predict unknown forces in dynamic morphing wing flights. 

Figure~\ref{fig:position} shows the position tracking (X, Y, Z) of the Aerobat during simulated open-loop flight scenarios. The X-coordinate indicates consistent forward motion, while the Y-coordinate shows a steady shift in the negative direction, reflecting lateral movement. The Z-coordinate oscillates slightly, capturing the vertical motion due to wing flapping, which validates the model’s ability to simulate complex, multi-directional flight paths in an open-loop control framework. Additionally, Fig.~\ref{fig:stick-diagram}, the stitch diagram, provides a visual representation of the Aerobat’s top-view trajectory over time.

Figure~\ref{fig:states} illustrates key aspects of the Aerobat's dynamics, including joint angles, body orientation, and linear velocity. The shoulder and elbow joint angles exhibit periodic motion, crucial for the flapping mechanism, while the Euler angles (roll, pitch, yaw) remain stable with minor fluctuations, confirming the model’s effectiveness in simulating dynamic wing morphing. The body’s linear velocities (X, Y, Z) show smooth, periodic trajectories that correspond with the flapping cycles, validating the model’s ability to predict temporal velocity changes resulting from aerodynamic forces and wing movements.

In this detailed analysis of force estimation for the Aerobat's wing, Fig.~\ref{fig:estim-diagram} exemplifies the efficacy of the conjugate momentum-based observer in precisely estimating unknown external forces in the inertial frame.

The top panel (portion of the Fy component), marked by the blue boxed region, illustrates the Gaussian noise integrated into the force model to simulate sensor noise and environmental variability. This noise, defined by Eq.~\eqref{eq:noise}, challenges the observer's ability to maintain accurate force estimation under realistic conditions. The close alignment in the graph, where the estimated forces (red line) match the actual forces (black line), highlights the observer's precision and reliability in differentiating signal from noise in noise-prone environments.

The magenta-boxed regions illustrate the observer's response to a sudden step input introduced between 1 and 1.6 seconds. This step input simulates an unexpected external force, such as a gust or obstacle impact, which is essential for evaluating the observer's responsiveness and adaptability. The graph shows that the estimated forces promptly adjust to the step change, closely mirroring the actual force profile. This demonstrates the observer's ability to accurately predict abrupt force changes and its potential to enhance the Aerobat’s adaptive responses in dynamic and unpredictable flight conditions.

To quantitatively evaluate the model’s performance, statistical metrics were applied to measure the accuracy of force predictions. Specifically, \( R^2 \) values were calculated for each force component: \( R^2_{Fx} = 0.7448 \), \( R^2_{Fy} = 0.9970 \), and \( R^2_{Fz} = 0.9991 \). These metrics demonstrate that the model predicts forces along the Y and Z axes with exceptional accuracy, capturing over 99\% of the variance. Furthermore, the prediction accuracy for the X-axis, while slightly lower, still explains 74.48\% of the variance. The lower $R^2$ value for the $F_x$ component could be attributed to factors such as continuous pitching motions of the Aerobat, as shown in Fig.~\ref{fig:states}, which induce greater variability in the force along this axis. This suggests that while the model performs strongly overall, further refinement may be needed to improve force estimation along the X-axis.

\section{Conclusion}
\label{sec:concl}
This paper presented the application of a Conjugate Momentum-based Observer for estimating unknown external forces acting on the Aerobat, a bio-inspired robotic platform with dynamically morphing wings. Our approach demonstrated the observer's effectiveness in accurately detecting external disturbances, such as wind gusts and unmodeled impacts, which are challenging to capture with traditional force estimation methods.

Through simulations, we validated the observer's ability to estimate these external forces under various flight conditions, including scenarios with Gaussian noise and sudden step inputs. The results show high accuracy in force estimation, particularly for the Y and Z components, with R² values exceeding 0.99. While the X component showed slightly lower accuracy (R² = 0.7448), possibly due to continuous pitching motions, the overall performance suggests significant potential for improving flight control in dynamic environments.

Future work will focus on integrating the observer into the closed-loop control and testing its performance in real-world flight experiments, with the goal of enabling more robust and autonomous navigation for bio-inspired robotic systems. This integration has the potential to significantly enhance the Aerobat's ability to adapt to unpredictable external forces, paving the way for more versatile and resilient flying robots.

\printbibliography

\end{document}